\documentclass[pdflatex,sn-mathphys-num]{sn-jnl}

\usepackage{graphicx}
\usepackage{multirow}
\usepackage{amsmath,amssymb,amsfonts}
\usepackage{amsthm}
\usepackage{mathrsfs}
\usepackage[title]{appendix}
\usepackage{xcolor}
\usepackage{textcomp}
\usepackage{manyfoot}
\usepackage{booktabs}
\usepackage{algorithm}
\usepackage{algorithmicx}
\usepackage{algpseudocode}
\usepackage{listings}

\usepackage{siunitx}
\usepackage{stfloats}
\usepackage{float}
\usepackage{url}
\usepackage{bm}

\theoremstyle{thmstyleone}

\theoremstyle{thmstyletwo}

\theoremstyle{thmstylethree}

\begin{document}

\title[Article Title]{Deriving Decision-Relevant Visual Features from Human Gameplay for Reinforcement Learning}

\author*[1]{\fnm{Henrik} \sur{Krauss}}\email{henrik1.krauss@gmail.com}
\author[2]{\fnm{Takehisa} \sur{Yairi}}\email{yairi@g.ecc.u-tokyo.ac.jp}

\affil*[1]{\orgdiv{Department of Advanced Interdisciplinary Studies}, \orgname{The University of Tokyo}, \orgaddress{\city{Tokyo}, \postcode{153-8904}, \country{Japan}}}

\affil[2]{\orgdiv{Research Center for Advanced Science and Technology}, \orgname{The University of Tokyo}, \orgaddress{\city{Tokyo}, \postcode{153-8904}, \country{Japan}}}

\abstract{This study introduces a method for deriving decision-relevant visual features from human gameplay alone, leveraging offline attention techniques from reinforcement learning (RL). We propose contextualized, task-relevant (CTR) attention networks, which generate spatial attention maps from both human and RL agent gameplay in Atari environments. We evaluate the maps by action prediction under masking, by alignment with a temporally integrated overt attention (TIOA) model trained on eye-tracking data, and against a center-prior baseline. Human CTR maps are more sparse than the matched agent maps and align more strongly with TIOA. A qualitative comparison further suggests that these decision-relevant features line up well with what a player's internal attention may contain (overt gaze, together with decision-relevant regions of covert attention and short-term memory), while no label for such internal attention exists. As a further application, we use the CTR attention maps to guide RL agents. At 16\% mean activation, CTR-masked agents reach a normalized relative score of $1.10$ against a strong full-observation baseline, a minor improvement that is not statistically significant (paired $p=0.238$), and significantly outperform TIOA-masked and inverted-mask agents. This work provides an analysis framework based on gameplay data only to obtain such features for cognitive analysis and RL.}


\maketitle

\section{Introduction}\label{sec1}

\label{sec:introduction}
\begin{figure}[t]
    \centering
    \includegraphics[width=0.5\linewidth, trim=0 35 0 0, clip]{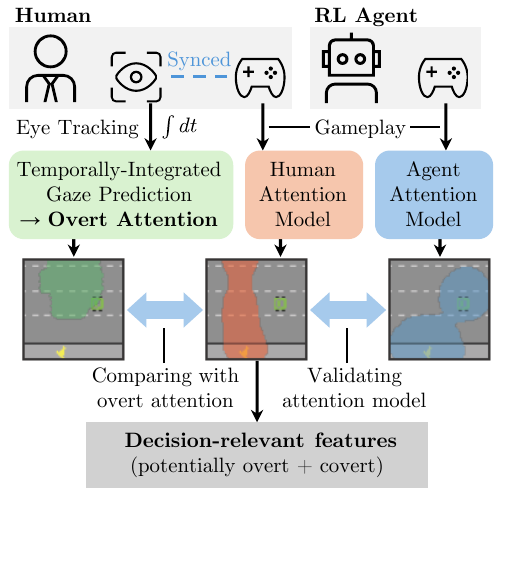}
    \caption{The goal of this study is to derive decision-relevant visual features solely from human and agent gameplay analysis. An attention model extracts contextually task-relevant features from human and reinforcement learning (RL) agent gameplay.
    A temporally integrated gaze prediction model (TIOA) is used as a point of comparison for alignment with overt attention.}
    \label{fig:intro}
\end{figure}

The frame problem~\cite{mccarthy1981some} represents an ongoing challenge in artificial intelligence (AI) research: How does an agent determine what information is relevant in complex, ever-changing environments? Understanding how humans allocate attention in task-based scenarios can provide valuable insights into this problem to eventually  design more intelligent, autonomous agents.
By integrating research on human attention with visual attention-based reinforcement learning (RL), we can gain a deeper understanding of attention’s role in human task learning while also improving the design of intelligent, autonomous agents.

As visual attention methods become more common in deep learning~\cite{hassanin2024visual}, several recent studies have explored incorporating attention mechanisms including human attention information into RL. These methods aim to either make an agent’s actions more interpretable, contributing to explainable AI, or improve learning efficiency and final performance. The attention approaches can be classified as offline-trainable and online-trainable.
In offline-trainable attention methods such as~\cite{greydanus2018visualizing,manchin2019reinforcement,gopalakrishnan2020unsupervised, wu2021self, shi2020self,nikulin2019free, ma2025don,chen2024focus}, an attention or saliency map is derived in a self-supervised way—either from the state (i.e., visual input) or from state-action-reward transitions (i.e., experience replay).
This map is then used for pre-processing the input image or features by weighting or selection before processing in the policy network, or can even be applied after training for interpretation purposes too.
In online methods such as~\cite{shang2023active, mnih2014recurrent, kulkarni2019unsupervised, mott2019towards}, attention is integrated directly into the learning process, for example, by additional attentive actions assigned to the agent, as in active vision reinforcement learning~\cite{shang2023active}.
These attention mechanisms have been widely used in visual deep RL for Atari games but have also been applied to robotic navigation~\cite{salter2021attention} and manipulation tasks~\cite{james2022coarse}.

Apart from applying an agent's own attention map, agents can also be augmented with human attention information. Zhang et al.~\cite{zhang2019leveraging} provide a comprehensive survey of methods for leveraging human guidance in deep RL, categorizing attention-based learning as one approach for agents to learn from humans. This has been shown to improve learning performance in imitation learning (IL), including behavioral cloning~\cite{zhang2020atari} and inverse RL~\cite{zhang2018agil,saran2021efficiently}.
Human attention information can be obtained through recording eye-tracking data during gameplay~\cite{zhang2020atari}, and predicted using gaze prediction models~\cite{borji2014what,yang2020predicting}. However, as Zhang et al.~\cite{zhang2019leveraging} note, existing work in this area has primarily focused on using gaze (overt attention) as the sole attention signal.
There is also research on task-oriented human gaze and attention prediction.
In~\cite{borji2014what}, a top-down gaze prediction model is built using a dynamic Bayesian network to predict human gaze positions during gameplay.
Yang et al.~\cite{yang2020predicting} used inverse reinforcement learning to model task-oriented attention in visual search paths on images.
However, to fully understand human decision making in task-based scenarios, actual or predicted gaze data alone is not sufficient. This is because \emph{gaze data only reveals overt attention}~\cite{posner1980orienting}, i.e., the attention visible to an observer. Covert attention is the mental focus on something without necessarily looking at it—such as an object in the visual periphery or in memory. Because it involves no observable movements, it can not be measured by eye tracking and is challenging to model~\cite{borji2014what, guo2021machine,leong2017dynamic}.

Guo et al.~\cite{guo2021machine} conducted an in-depth comparison of human gaze prediction with RL agents' perturbation-based attention (saliency map), showing that agent attention becomes more similar to human gaze when the discount factor is reduced from the typical 0.99 (long-term reward focus), and that agents fail more often when their attention diverges from human gaze. However, their analysis relied exclusively on gaze (overt attention) and therefore did not account for covert attention or short-term memory.
This study points out the limitations of using gaze data (overt attention) only to study human decision making and to build better human attention-augmented agents. 
\emph{If we had access to the visual features that humans use for decisions, including both decision-relevant overt and covert attention, they could be used to further improve intelligent, autonomous agent design}.
Without the need for additional brain activity data (such as in~\cite{leong2017dynamic}), it remains to be explored whether such decision-relevant features can be derived from gameplay alone.

In this paper, we define \emph{decision-relevant features} as the visual features of the current state that are sufficient for predicting the player's action. The CTR attention map is the spatial mask over those features. Related offline attention and saliency methods (e.g., Greydanus et al.~\cite{greydanus2018visualizing}, Wu et al.~\cite{wu2021self}, Shi et al.~\cite{shi2020self}, Nikulin et al.~\cite{nikulin2019free}, and the agent maps studied by Guo et al.~\cite{guo2021machine}) typically explain or sparsify an \emph{agent} policy. Our focus is on maps trained to predict \emph{human} actions and compared with human gaze. We discuss carefully whether these features also line up with \emph{internal attention}, the combination of overt attention (gaze), covert attention, and short-term visual memory that a player would use in a given state, as an interpretation of the maps, not as a quantity we measure directly, as no true label exists to confirm this.

The two main contributions of this study are the following: (1) We derive decision-relevant visual features from gameplay. The approach of this study is visualized in Fig.~\ref{fig:intro}. (1a) We propose a contextualized, task-relevant (CTR) attention network trained hand-in-hand with an action prediction network, inspired by offline attention methods in reinforcement learning (similar to visual attention in behavioral cloning~\cite{mohandoss2019visual}). Applied to the gameplay (experience replay) of humans and RL agents playing Atari games, this network selects features relevant to the current state at various sparsity levels. (1b) We compare these attention maps against a custom temporally-integrated overt attention (TIOA) network, trained on eye-tracking data from the same human Atari gameplay, and against a center-prior baseline. Through this analysis we assess how well the derived features align with gaze (TIOA) and discuss their relation to internal attention. (2) We apply the derived features to reinforcement learning by masking the encoded observation of RL agents, and evaluate whether they are valuable and sufficient for learning relative to a strong full-observation baseline, an inverted mask, and a TIOA mask.

In Sec.~\ref{sec:methods}, firstly the experimental setup and used datasets are explained and the architecture of the CTR attention and TIOA network as well as the temporal integration of eye-tracking data explained.
The evaluation is split in three sections, where in Sec.~\ref{sec:evaluation_a} the action predictor is evaluated, in Sec.~\ref{sec:evaluation_b} the revealed attention maps are analyzed quantitatively and qualitatively, and in Sec.~\ref{sec:evaluation_c} the application of attention to reinforcement learning is presented.
The results are discussed and concluded in section~\ref{sec:discussion}.

\section{Methods}
\label{sec:methods}
\begin{figure*}
    \centering
    \includegraphics[width=\linewidth, trim=0 65 0 0, clip]{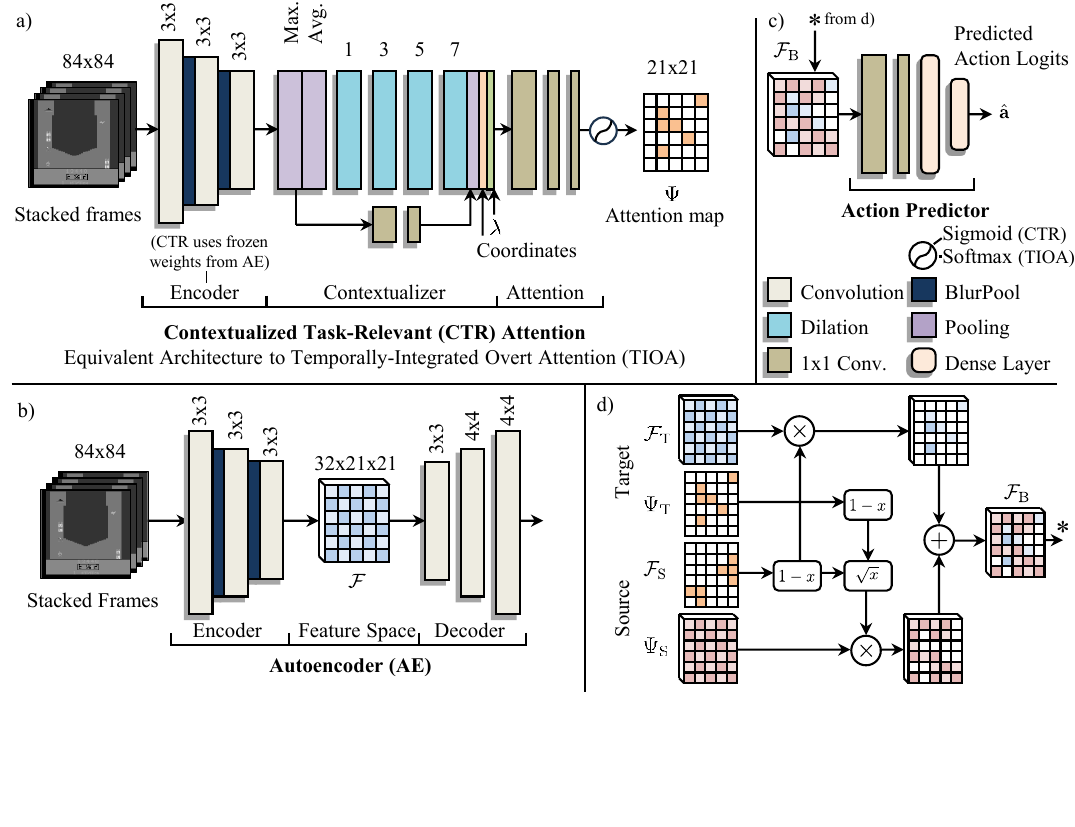}
    \\[-.5ex]
    \noindent\rule{\linewidth}{0.8pt}
    \begin{minipage}{0.8\linewidth}
        \begin{picture}(0,0)
            \put(-12,50){%
                {\fontfamily{ptm}\fontsize{8}{8}\selectfont e)}%
            }
        \end{picture}
        \includegraphics[width=0.97\linewidth]{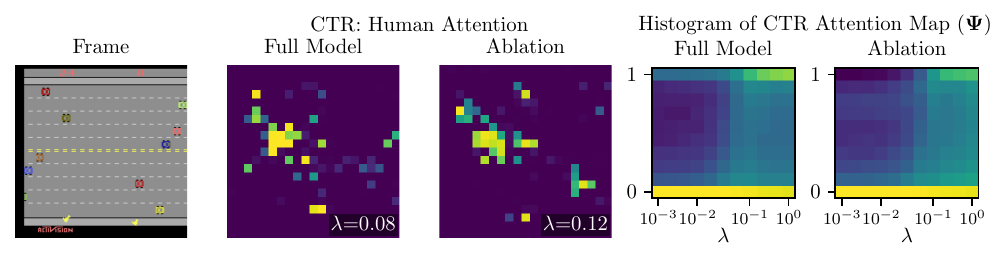}
    \end{minipage}
    \caption{
    a) Contextualized task-relevant (CTR) attention network as well as temporally-integrated overt attention (TIOA) network architecture;
    b) Autoencoder (AE) architecture;
    c) Action predictor architecture;
    d) Method of creating a blended feature space $\mathcal{F}_\mathrm{B}$ from a target and source for the action predictor;
    e) Example frame and attention map (left), and log$_{10}$ histogram of activations (right) over different sparsity control factors $\lambda$ for the human CTR attention network with and without feature blending. Feature blending drives the activation distribution toward binarization (values near 0 or 1).
    }
    \label{fig:architecture}
\end{figure*}

\begin{figure}[!htb]
    \centering
    \includegraphics[width=0.5\linewidth, trim=0 135 0 0, clip]{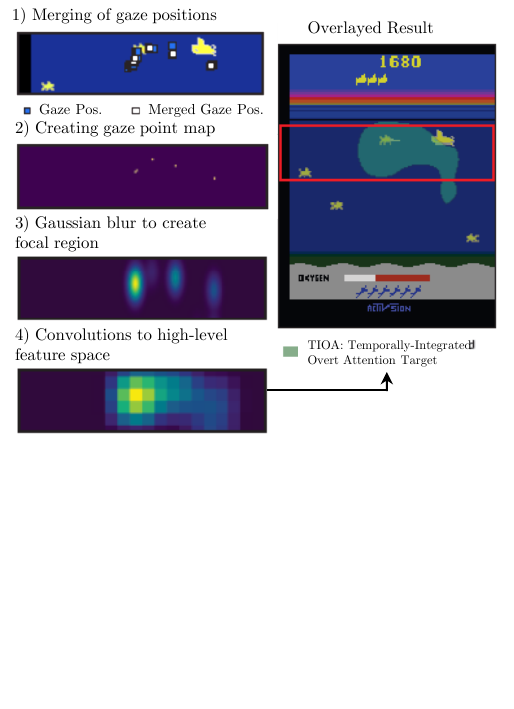}
    \caption{Step-by-step construction of a temporally-integrated overt attention target map from gaze positions for training of the TIOA network.}
    \label{fig:gaze_target}
\end{figure}

\subsection{Study Overview and Datasets}
The study plan illustrated in Fig.~\ref{fig:intro} will be implemented using Atari gameplay data from both humans and reinforcement learning agents, including human eye-tracking information. Atari games are chosen as a proven research platform for visual attention-based reinforcement learning.
Their limited exploratory freedom further ensures that humans and agents encounter similar states despite differences in performance, making them well-suited for this comparison.
Six Atari games (Freeway, MsPacman, Enduro, Seaquest, SpaceInvaders, Riverraid) are chosen to cover a variety of game types, visuals, and action counts.

For agent gameplay, we use pre-trained agents from~\cite{gogianu2022agents}, which are based on Munchausen Deep Q-Networks (M-DQN)~\cite{vieillard2020munchausen}. These agents outperform humans on average, achieving an inter-quartile mean score of approximately 1.8. A replay memory is generated for each game using a frame skip of four, meaning each action is repeated four times per transition.
We use the Atari-HEAD dataset~\cite{zhang2020atari}, which contains amateur human gameplay of the same Atari games along with simultaneous eye-tracking data. Human subjects played in a semi-frame-by-frame mode at a maximum of \qty{20}{Hz}, mitigating potential state-action mismatches caused by reaction time delays. The state-action transitions are converted into an experience replay memory similar to the agent's but extended with the last gaze position for each frame. All available gameplay from four human subjects is included, but only every fourth frame is used to ensure consistency with agent transitions, resulting in approximately 3–6 hours of actual gameplay per game. The agent experience replay is then adjusted to match the total length of the available human gameplay data\footnote{Replay memory sizes are: $\num{1.0e5}$ (MsPacman, Seaquest, Riverraid), $\num{1.2e5}$ (Enduro, SpaceInvaders), $\num{0.5e5}$ (Freeway).}.

Although four subjects were recorded for each game, the available data per individual is deemed insufficient for robust subject-specific analysis. We therefore aggregate the gameplay data from all subjects to study general differences between human and agent maps.
For each game, and for both agent and human replay memory, we train one attention model (CTR attention) and one human gaze predictor (TIOA model). Training the attention network also involves an autoencoder and an action predictor. To prevent overreliance on elements apart from the main scene, interface elements and score indicators are deactivated. The neural network models are implemented and trained in PyTorch, with replay memory randomly split into 80\% training and 20\% validation \emph{transitions} (not episodes or subjects). Because inputs use a four-frame stack and the TIOA target integrates about $3\,\mathrm{s}$ of gaze, frames near a split boundary can contribute to both sides. ``Unseen states'' therefore refers to held-out transitions from the same trajectories and the same four pooled players, and reported action prediction accuracies can be optimistic relative to a new episode or player. All models are trained for at least 300 epochs until convergence.%
\footnote{The code is made publicly available at: \url{https://github.com/UThenrik/revealing-attention}}

\subsection{Contextualized, Task-Relevant (CTR) Attention Network}
\label{subsec:CTR}
We propose a contextualized, task-relevant (CTR) attention network, visualized in Fig.~\ref{fig:architecture}(a), which predicts an attention map
\begin{equation}
    \bm{\Psi} = f_\mathrm{CTR}(\bm{s}, \lambda, \theta) \in \mathbb{R}^{1 \times 21 \times 21}
\end{equation}
based on the current state $\bm{s}\in \mathbb{R}^{4 \times 84 \times 84}$ and a sparsity controlling value $\lambda \in [0, 1]$, parameterized by $\theta$. The state is represented as a stack of four game frames, each a grayscale image of size $84 \times 84$, downscaled from an RGB image of size $3 \times 210 \times 160$ as commonly done in Atari RL research. The state is first processed through a three-stage convolutional encoder, similar to the architecture in~\cite{wu2021self}, reducing spatial resolution fourfold. The encoder weights are initialized from a pre-trained autoencoder, as illustrated in Fig.~\ref{fig:architecture}(b) and described in the next paragraph.
After encoding, the features are contextualized using four serial dilated convolution layers with increasing dilation rates, concatenated with global maximum and average pooling channels, total spatial coordinates (similar to CoordConv~\cite{liu2018intriguing}), and the expanded attention sparsity value $\lambda$. The final attention map is generated through three serial pointwise convolution layers, outputting a single-channel map of size $1 \times 21 \times 21$. Leaky ReLU activations are applied between layers, with a sigmoid activation at the output to ensure values remain in the $[0,1]$ range.

For training the CTR attention network, we use a pre-trained autoencoder (AE), as illustrated in Fig.~\ref{fig:architecture}(b). This AE is trained for each game separately on states from both agent and human replay memory and encodes the state into a feature tensor $\bm{\mathrm{F}}\in \mathbb{R}^{32 \times 21 \times 21}$ with 32 feature channels and spatial dimensions matching $\bm{\Psi}$.
After the first two encoder layers we apply BlurPooling~\cite{pmlr-v97-zhang19a} with stride of 2 to ensure the feature space is shift-invariant while allowing for fourfold reduction of spatial resolution.

By applying the attention map at this encoded level, we achieve a balanced mix of spatial, feature-based, and object-based attention. For example, a single feature can preserve spatial location while also encoding an object (e.g., a car’s position in Freeway) and/or a higher-level representation (e.g., the car’s driving direction).
The CTR attention network is trained alongside an action predictor, whose architecture is shown in Fig.~\ref{fig:architecture}(c). We architecturally separate the attention network from the action predictor. The CTR network produces masks that are applied as an explicit input before the action predictor. Both networks are trained jointly with the action-prediction loss (the mask itself is not detached), so the separation makes the mask an interpretable intermediate rather than guaranteeing that the maps are free of optimization artifacts. The action predictor estimates the action for the current target state based on a feature space:
\begin{equation}
    \hat{\bm{a}} = f_\mathrm{AP}(\bm{\mathrm{F}}, \theta)
\end{equation}
where $f_\mathrm{AP}$ denotes the action predictor network and $\bm{\mathrm{F}}$ is a feature tensor, which can be the blended feature space, an attention-masked feature space, or the raw feature space.
\begin{equation}
\bm{\mathrm{F}}_\mathrm{B} = \bm{\mathrm{F}}_\mathrm{T} \odot \bm{\Psi}_\mathrm{T} + 
\bm{\mathrm{F}}_\mathrm{S} \odot \big( (1 - \bm{\Psi}_\mathrm{T}) \odot (1 - \bm{\Psi}_\mathrm{S}) \big)^{\circ \frac{1}{2}}
\end{equation}
This blended space overlays the target feature tensor $\bm{\mathrm{F}}_\mathrm{T}$ and the source feature tensor $\bm{\mathrm{F}}_\mathrm{S}$, which is derived from a randomly sampled state. Here, $ \odot$ and $\circ$ denote element-wise multiplication and exponentiation, respectively. The blending process for the feature spaces is visualized in Fig.~\ref{fig:architecture}(d). This feature blending mechanism drives the CTR attention network toward binarized attention maps. By requiring the network to distinguish between target and source features while maintaining action prediction accuracy, attention values are encouraged to converge toward either 0 (irrelevant) or 1 (relevant). An ablation study in Fig.~\ref{fig:architecture}(e) shows that without feature blending, the network produces less binarized activation distributions.

The total loss for joint training of the CTR attention network and the action predictor is
\begin{align}
\label{eq:total_loss}
\mathcal{L}_\mathrm{total} =\;& \mathcal{L}_\mathrm{AP,B} + \mathcal{L}_\mathrm{AP} \\
&+ \beta\, \mathcal{L}_{\lambda,1} + (1-\beta)\, \mathcal{L}_{\lambda,2}
\end{align}
where
\begin{align}
\mathcal{L}_{\lambda,1} &= \frac{1}{n_\lambda} \sum_{i=1}^{n_\lambda} \frac{|\overline{\bm{\Psi}}_i - \lambda_i|}{|\lambda_i| + \epsilon} \\
\mathcal{L}_{\lambda,2} &= \mathcal{L}_{\lambda,1} \cdot \mathbb{I}(\overline{\bm{\Psi}}_i > \lambda_i)
\end{align}
and $\beta = \max(0, 1 - \frac{j_\mathrm{epoch}}{100})$ controls loss weighting over training epochs. Here, $n_\lambda$ is the number of sampled $\lambda$ values per batch (we use $n_\lambda=6$), $\overline{\bm{\Psi}}_i$ is the mean of the attention map for sample $i$, $\lambda_i$ is the corresponding sparsity target, $\epsilon$~is a small constant for numerical stability, and $\mathbb{I}(\cdot)$ is the indicator function.
We use the cross-entropy
\begin{equation}
\mathrm{CE}(\bm{a}, \hat{\bm{a}}) = -\frac{1}{n_\mathrm{b}} \sum_{j=1}^{n_\mathrm{b}} \sum_{i=1}^{n_\mathrm{a}} a_{i,j} \log(\hat{a}_{i,j})
\end{equation}
where $\bm{a}$ is the true action (one-hot encoded), $\hat{\bm{a}}$ is the predicted action probability, $n_\mathrm{b}$ is the batch size, and $n_\mathrm{a}$ is the number of actions.
The action prediction losses are then given by
\begin{align}
\mathcal{L}_\mathrm{AP,B} &= \mathrm{CE}(\bm{a}, \hat{\bm{a}}^\mathrm{B}) \\
\mathcal{L}_\mathrm{AP}   &= \mathrm{CE}(\bm{a}, \hat{\bm{a}})
\end{align}
where $\hat{\bm{a}}^\mathrm{B}$ and $\hat{\bm{a}}$ are the predicted action probabilities from the blended and full feature spaces, respectively.
The attention maps used for the blended feature space action prediction loss $\mathcal{L}_\mathrm{AP,B}$ are the same as those used for the sparsity losses $\mathcal{L}_{\lambda,1}$ and $\mathcal{L}_{\lambda,2}$, which is why concurrent training of the CTR attention network and action predictor is required.

The sparsity-controlling factor $\lambda$ restricts the average activation of the attention map, specifying the maximum ratio of features that can be active. The loss term $\mathcal{L}_{\lambda,2}$ enforces this. While we assume that more features may improve action prediction by providing more information, we want to avoid unnecessary activations. The auxiliary loss $\mathcal{L}_{\lambda,1}$ also controlling minimum sparsity in early training is introduced for training stabilization. The sparsity factor $\lambda$ is implemented as a controllable network input, enabling post-training adjustment across multiple sparsity levels essential for our comparative analysis in Sec.~\ref{sec:evaluation_a}. Without regularization, attention maps converge to full activation to maximize information, defeating the purpose of selective attention.

The total loss~\eqref{eq:total_loss} combines cross-entropy terms, ensuring the action predictor selects the correct action, with a sparsity regularization term that keeps the attention map controllable. The blended feature space encourages the CTR attention network to focus on contextually task-relevant features from the target and suppress irrelevant features from the source, driving $\bm{\Psi}$ values toward $0$ or $1$. As a result, the CTR attention network learns to spatially \emph{extract features relevant to the player's decision in each gameplay context}, adapting flexibly to different sparsity levels specified by $\lambda$. For example, the network is trained such that, if we input $\lambda=0.05$, it activates on average at most 5\% of the features.

The action predictor is trained using a learning rate of $1 \times 10^{-3}$, while the CTR attention network is trained with half that rate, i.e., $5 \times 10^{-4}$, which encourages stable joint optimization.
After training, the CTR attention models can be used independently of the autoencoder and action predictor.

It is noted that the CTR attention network can be used plug-and-play with the self-supervised attention network from~\cite{wu2021self}, which has been shown to improve the rate of convergence and performance of deep RL methods in Atari games, while additionally offering task-relevant and contextualized feature selection. This will therefore be investigated in Sec.~\ref{sec:evaluation_c}.
It could also be trained using inverse dynamics prediction, i.e. predicting an action $a_k$ from the current state $\bm{s}_k$ and the next $\bm{s}_{k+1}$. This approach would extract all features that are evidence of an action and could be more robust for applications in RL. However, we use forward action prediction as it directly extracts features that preceded and likely influenced the player's decision, aligning with our goal to reveal decision-relevant attention patterns.

\subsection{Temporally-Integrated Overt Attention (TIOA) Network}
In order to compare CTR maps with overt attention, a gaze-based point of comparison is needed. Rather than using only the instantaneous gaze positions (eye-tracking data) from the Atari-HEAD dataset, an integration of recent gaze positions is preferable, as it more closely mimics the temporal integration of visual sensory data in humans~\cite{fairhall2014temporal}, and therefore provides a stronger overt-attention baseline when discussing short-term visual memory.
To further enhance generalization to unseen states (and those of the agent), we propose a temporally-integrated overt attention (TIOA) network.
It features the same base architecture as the CTR attention network in Fig.~\ref{fig:architecture}(a) and therefore predicts temporally-integrated overt attention on the fourfold resolution-reduced feature space. Its output shall denote a probability distribution and therefore a softmax function is used after the final layer.

The TIOA network is trained to predict a temporally integrated overt attention map, denoted as $\hat{\bm{\Gamma}}$. For training, we construct a target map $\bm{\Gamma}^*$ by weighting and integrating recent gaze positions recorded in the Atari-HEAD dataset. The step-by-step construction of $\bm{\Gamma}^*$ is visualized in Fig.~\ref{fig:gaze_target}.
Human visual integration for complex visual sequences is considered to be around $2$--$\qty{3}{s}$, with a steep increase in difficulty to integrate information around $\qty{2.5}{s}$~\cite{fairhall2014temporal}. As the human subjects played the games at a maximum frequency of $\qty{20}{Hz}$, in step~1), we take the most recent $60$ gaze positions and merge very close points within $\qty{6}{px}$, as they are expected to correspond to the same object in the Atari game. In step~2), a gaze point map is created where the individual points are weighted by a negative exponential function
\begin{equation}
    w(\tau)=1.1-10^{\frac{\tau}{c}-1}
\end{equation}
where $\tau$ denotes the time lag (in seconds) relative to the current time, i.e., $\tau=0$ for the most recent gaze point and increasingly negative values for earlier gaze points. Here, $c=-2.95$. This weighting emphasizes more recent gaze positions, resulting in the most recent one being weighted $w(\tau=0)=1$ and the last considered one by $w(\tau=-\qty{2.95}{s})=0.1$. The weighting function is modeled after the difficulty to follow scrambled video sequences of varying lengths reported in~\cite{fairhall2014temporal}. In step~3), Gaussian blur is applied with a standard deviation of one visual degree to obtain a distribution representing the focal region~\cite{le2013methods,zhang2020atari}. This appears non-symmetric as the games were stretched to wide-screen in the Atari-HEAD experiment. Finally, in step~4), the blurred map is reduced to the same resolution as the CTR attention maps by applying three convolutions. These convolutions use fixed kernels, set to 1, with the same kernel size and stride as those in the CTR, TIOA, and AE encoder. The target for the TIOA is finalized by normalizing it to $\sum_{i}^{n_\mathrm{f}}\bm{\Gamma}^*=1$.
The TIOA is trained using a Kullback-Leibler divergence loss
\begin{equation}
    \mathcal{L}_\mathrm{TIOA} = \frac{1}{n_\mathrm{b}} \sum_{j=1}^{n_\mathrm{b}} \sum_{i}^{n_\mathrm{f}} \hat{\bm{\Gamma}}_{i,j} \log \frac{\hat{\bm{\Gamma}}_{i,j}}{\bm{\Gamma}_{i,j}^*}
\end{equation}
between the temporally integrated gaze target map $\bm{\Gamma}^*$ and its prediction $\hat{\bm{\Gamma}}$ with the number of spatial features ${n_\mathrm{f}}=441$.

\section{Evaluation I: Action Prediction under Attention Masking}
\label{sec:evaluation_a}
\begin{figure*}
    \centering
    \includegraphics[width=\linewidth,trim=0 0 0 0, clip]{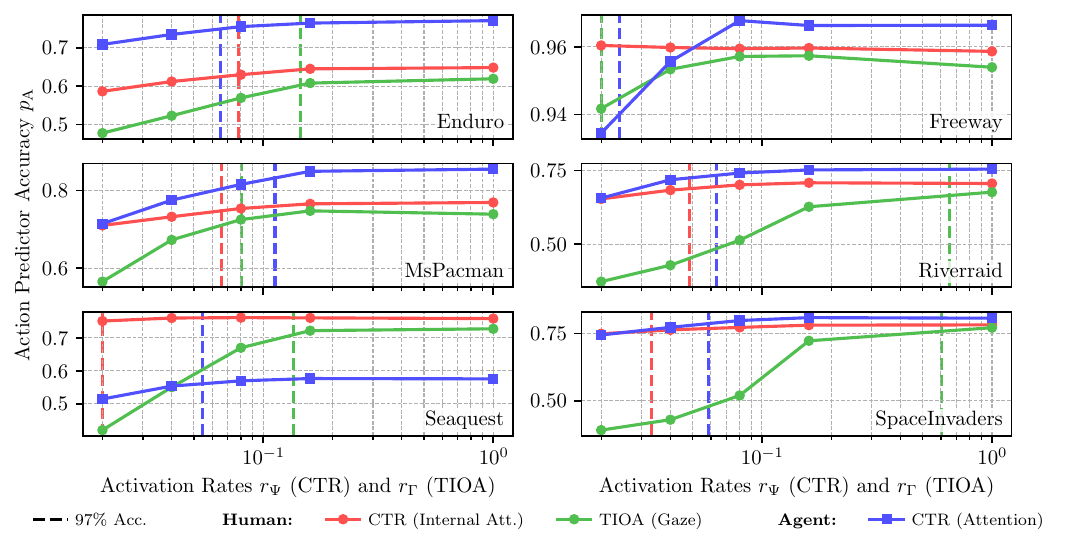}
    \vspace{-4 mm}
    \caption{Action predictor accuracies over unseen attention-masked states for human (CTR and TIOA) and agent (only CTR) gameplay at different attention activation rates/ sparsities. Dashed lines indicate the minimum activation rate where 97\% of the maximum action prediction accuracy is reached.}
    \label{fig:ap_accuracy}
\end{figure*}

\begin{figure}[!htb]
    \centering
    \includegraphics[width=0.5\linewidth]{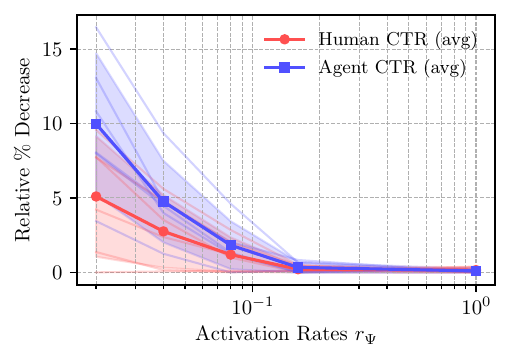}
    \caption{Relative percentage decrease in action prediction accuracy from maximum comparing human CTR vs. agent CTR across different sparsity levels. Mean and 95\% confidence interval across six games. Agent attention shows approximately twice the relative accuracy decrease compared to human attention, supporting that agents have more distributed attention.}
    \label{fig:ap_rel_decrease}
\end{figure}

In this section, we evaluate the action prediction accuracy for both human and agent gameplay under full observation and under attention masking. For the human action prediction, we assess attention masking using both CTR and TIOA attention maps of varying sparsity that is controlled by the sparsity value~$\lambda$. For the agent, we evaluate attention masking using only the CTR attention maps. The action predictors used here are trained directly on the attention-masked states produced by the trained CTR and TIOA networks.
Fig.~\ref{fig:ap_accuracy} shows the resulting action predictor accuracies
\begin{equation}
p_\mathrm{A} = \frac{1}{n_\mathrm{s}} \sum_{j=1}^{n_\mathrm{s}} 
\mathbb{I}\Big(\arg\max_i \hat{a}_{i,j} = \arg\max_i a_{i,j}\Big)
\end{equation}
over all unseen states $n_\mathrm{s}$ for human and agent gameplay, with activation rates $r_{\Psi}$ and $r_{\Gamma}$ evaluated at $0.02$, $0.04$, $0.08$, $0.16$, and $1.0$.
For the CTR attention maps, the activation rate $r_{\Psi}$ is achieved by controlling $\lambda$ so that as the average activation in the unseen states results to:
\begin{equation}
r_{\Psi} \approx \frac{1}{n_\mathrm{s}} \sum_{i=1}^{n_\mathrm{s}} \overline{\Psi}_i
\end{equation}
For TIOA, which produces a probability map, we binarize the attention map by selecting the top $r_{\Gamma}$ fraction of features (i.e., those with the highest logits), setting them to $1$ and the rest to $0$:
\begin{equation}
\hat{\bm{\Gamma}} = \mathbb{I}\left(\hat{\Gamma} \geq \tau_k\right)
\end{equation}
where $\tau_k$ is the threshold such that $r_{\Gamma}$ fraction of features are activated.

We observe that accuracies for the human actions are higher than in the original Atari-HEAD study~\cite{zhang2020atari}, likely as we are predicting based on frame-stacked states.

In total, four main findings can be observed. Firstly, higher activation rates, i.e., more information leads to more accurate action prediction.
This is the case for all games except Freeway.
Secondly, the CTR attention maps are significantly more efficient than the gaze data-based TIOA maps, achieving higher action prediction accuracies with fewer activated features. This is expected, because CTR is trained for action prediction whereas TIOA is a fixed top-$k$ over a gaze map. At a 16\% activation rate (as used in the RL experiments in Sec.~\ref{sec:evaluation_c}), a one-sided paired $t$-test across the six games yields $p=0.0093$ (mean accuracy difference $0.039$, 95\% CI $[0.010,\,0.069]$, paired effect size $d_z=1.40$). We report this as a property of the action-trained mask, not as evidence that covert attention is required.
We also note that the CTR network reads the full frame stack before emitting the mask, so locations that are later zeroed can still influence which features are retained. High accuracy at low activation means the retained features are sufficient for a predictor trained on that mask, not that they were the only information used to choose the mask.
The dashed lines in Fig.~\ref{fig:ap_accuracy} indicate the minimum activation rate where 97\% of the maximum action prediction accuracy is reached.
Thirdly, human actions can be accurately predicted with fewer features than agent actions, consistent with the conclusion of Guo et al.~\cite{guo2021machine} from an analysis of human gaze data. Our results extend this finding to decision-relevant features extracted via behavioral cloning. Fig.~\ref{fig:ap_rel_decrease} quantifies this difference. Agent attention shows approximately twice the relative accuracy decrease from maximum compared to human attention, though with large confidence intervals reflecting game-dependent variation. Direct quantitative comparison to Guo et al.\ is limited, as their work compared agent saliency maps with human gaze peaks, whereas CTR selects a broader set of action-predictive features.
Lastly, despite the expected greater gameplay variety in the human replay memory (due to the involvement of four different human subjects), the prediction accuracies vary by game but are generally similar to those of a replay memory of a single trained agent.

The findings suggest that the CTR network correctly identifies contextually decision-relevant features. 
Whether these maps also line up with human internal attention is discussed in Sec.~\ref{sec:evaluation_b}.

\section{Evaluation II: Analysis of Revealed Attention Maps}
\label{sec:evaluation_b}
\subsection{Visual Analysis of Attention Maps}
\label{subsec:visual_qualitative}
\begin{figure*}
    \centering
    \includegraphics[width=\linewidth, trim=0 0 0 0, clip]{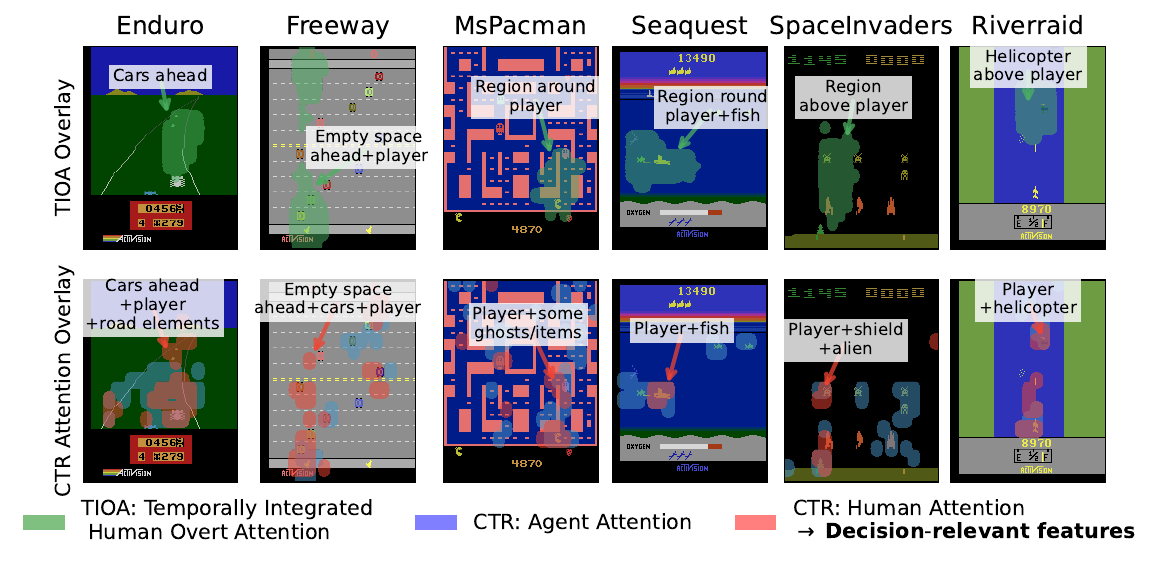}
    \vspace{-7 mm}
    \caption{Binarized attention map comparison: Example scenes (unseen states) with binarized TIOA (upper) and CTR attention map (lower) overlay. Human CTR maps more closely resemble the TIOA than the agent maps. They include parts of the TIOA map, nearby objects, and highly decision-relevant objects (see annotations), which we discuss as alignment with internal attention.}
    \label{fig:att_comparison}
\end{figure*}

\begin{figure}[!htb]
    \centering
    \includegraphics[width=0.5\linewidth, trim=0 0 0 0, clip]{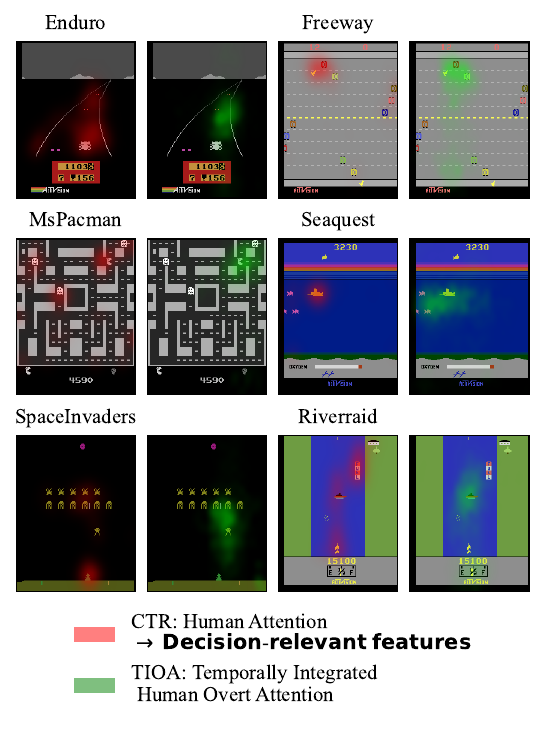}
    \vspace{-4 mm}
    \caption{Smooth attention map comparison: Example scenes (unseen states) with TIOA and Human CTR attention map that have been evaluated at multiple sparsity levels, smoothed and averaged. The MsPacman frame has been converted to grayscale for better visualization.}
    \label{fig:att_comparison_smooth}
\end{figure}

\begin{figure}[!htb]
    \centering
    \includegraphics[width=0.5\linewidth, trim=0 0 0 0, clip]{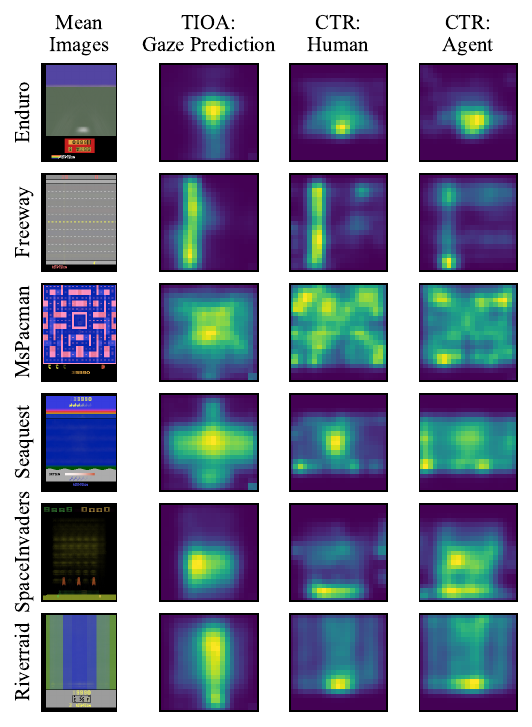}
    \caption{Heatmaps of TIOA and CTR attention maps over unseen states.}
    \label{fig:heatmaps}
\end{figure}

In Fig.~\ref{fig:att_comparison}, a scene is presented for each game with overlays of the TIOA and CTR maps. The selected scenes are intended to represent typical characteristics of the TIOA and CTR maps. Animations are provided in the supplemental materials.
To adjust for differences between human and agent attention patterns identified in Sec.~\ref{sec:evaluation_a} we base the chosen activation rates $r_\Psi$ off the 97\% of action prediction accuracy threshold from Fig.~\ref{fig:ap_accuracy}, with a lower limit of $0.02$ and an upper limit of $0.16$ resulting in the values shown in Tab.~\ref{tab:activation_rates}. The lower limit is the sparsest rate on the Evaluation I grid. For Freeway and Seaquest, human action-prediction accuracy is already at or above 97\% of its maximum at that rate, so those two corresponding activation rate values are clipped at $0.02$. In Fig.~\ref{fig:att_comparison} for better visualization, we use a stricter upper limit of $0.08$.
\begin{table}[!htb]
\centering
\setlength{\tabcolsep}{18pt}
\begin{tabular}{lcc}
\textbf{Game} & \textbf{Agent} & \textbf{Human} \\
\hline
Enduro        & 0.065  & 0.078 \\
Freeway       & 0.024 & 0.020  \\
MsPacman      & 0.112  & 0.066 \\
Seaquest      & 0.054 & 0.020  \\
SpaceInvaders & 0.059  & 0.033  \\
Riverraid     & 0.064  & 0.048 \\

\end{tabular}
\caption{CTR attention map activation rates $r_\psi$ for each game used in the analysis in Sec.~\ref{sec:evaluation_b}.}
\label{tab:activation_rates}
\end{table}
For the visualization of the CTR attention maps in Fig.~\ref{fig:att_comparison}, the maps are binarized at a threshold of $0.5$, and then upscaled. The TIOA map probabilities are binarized at a threshold in the range $[0.005,0.01]$ depending on the game\footnote{TIOA binarization thresholds are: $0.005$ (Freeway), $0.0075$ (Seaquest), $0.01$ (Enduro, MsPacman, SpaceInvaders, Riverraid).}.
\label{subsec:visualanalysis}
Providing additional scenes, Fig.~\ref{fig:att_comparison_smooth} shows the TIOA and Human CTR attention maps. The latter have been evaluated at multiple sparsity levels\footnote{The target sparsity levels used for the smoothed Human CTR maps are: $0.01$, $0.02$, $0.04$, and $0.08$.}, smoothed with a Gaussian kernel of standard deviation $1$ and averaged. This is to make use of the controllable sparsity to create a continuous attention map with similar distribution and appearance as the (un-binarized) TIOA maps. Lastly, Fig.~\ref{fig:heatmaps} presents overall heatmaps at the sparsities from Tab.~\ref{tab:activation_rates} over all unseen states (validation dataset).

It will be qualitatively analyzed for each game whether the TIOA and attention maps differ, and how the CTR maps relate to internal attention. The TIOA maps serve as a point of comparison for overt attention, as no true label for decision-relevant features or covert attention exists.
Generally, the TIOA maps are expected to overlap better with the human CTR attention maps but not fully. The non-overlapping part is where covert attention, short-term memory, or unattended yet action-predictive features would appear.

In Enduro, the TIOA predicts gaze toward the distant road and sometimes incoming cars, while the CTR similarly focuses on the cars and road boundaries, also incorporating the avatar's car. The human seems to attend further ahead, potentially due to slower reaction time than the agent. The avatar's car is a clear decision-critical object outside the gaze peak, available peripherally or from memory. This is the clearest single case for the internal-attention reading, while remaining compatible with an action-predictive feature the player may not attend. The heatmaps all show a focus towards the center.

In Freeway, the agent consistently attends to most cars and the player's avatar (a chicken crossing the road). In contrast, human CTR attention alternates between two distinct modes: when waiting, the focus is mostly on the avatar (seen in Fig.~\ref{fig:att_comparison_smooth}), and when crossing, attention shifts toward the road ahead and cars (seen in Fig.~\ref{fig:att_comparison}). The two modes can also be seen in quantitative analysis in the following Sec.~\ref{subsec:visual_quantitative}. The pattern aligns with human gaze behavior, where attention is primarily directed at the path ahead and objects near it. The human generally focuses stronger on the line the avatar is walking than the agent. This is predicted by the CTR network and can be seen clearly in the heatmap as well.

For MsPacman and Seaquest, the TIOA predicts human gaze primarily around the avatar character and a few opponents or collectibles. The human CTR map, especially in MsPacman, overlaps the gaze region and extends to nearby decision-critical objects. In the shown scene, the CTR attention map focuses on a relevant nearby ghost which is also looked at, but rather at the edge of the focus area. The agent's CTR map, on the other hand, extends attention to opponents and smaller items farther away. The agent's attention is stronger along the horizontal line along the avatar's vertical position. Notably, heatmaps show that both the human CTR and TIOA maps attend to various positions in the game environment.

In Space Invaders, the TIOA predicts gaze primarily on the attacking aliens rather than the avatar cannon. The agent CTR map, in contrast, distributes attention across both the avatar and most aliens. The human CTR map also focuses on the shields and slightly prioritizes the aliens immediately above the avatar while ignoring others, reflecting a more selective attention strategy.

In Riverraid, the TIOA predicts attention toward incoming targets and fuel fields ahead of the avatar's plane, primarily in the middle and upper third of the screen, as well as occasionally on the plane itself.
The human CTR map focuses on most dynamic objects, without clearly distinguishing between incoming opponents and purely aesthetic elements. However, in Fig.~\ref{fig:att_comparison_smooth} the attention on the aesthetic elements is much lower.
The agent's attention map also includes the river's edge, which marks an obstacle.
Riverraid is perhaps the most complex of the six games analyzed, as it features a constantly changing background and scenarios. The limited dataset size likely prevents the CTR network from extracting stable decision-relevant patterns from the human gameplay for this game.

\subsection{Similarity Analysis and Histograms of Attention Maps}
\label{subsec:visual_quantitative}
\begin{figure*}
    \centering
    \includegraphics[width=\linewidth, trim=0 0 0 0, clip]{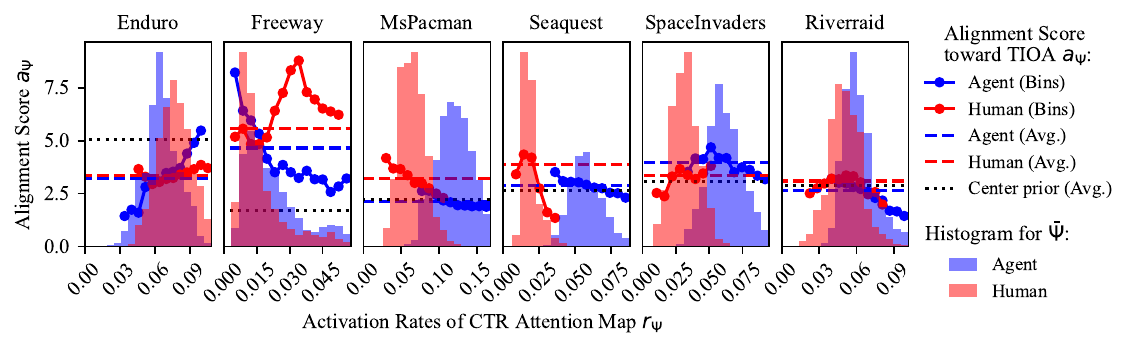}
    \vspace{-7 mm}
    \caption{Comparison for CTR agent and human attention map activation rates and alignment score toward the TIOA maps. CTR human attention maps have lower activation rates (higher sparsity) than the CTR agent maps and in average higher alignment scores (higher similarity) to the TIOA. The dotted line is the mean alignment of a binary center prior matched to the human CTR active count.}
    \label{fig:histograms}
\end{figure*}
An analysis of the characteristics of the CTR attention maps is performed. The CTR attention maps are evaluated for a specific attention activation rate as listed in Tab.~\ref{tab:activation_rates}. $1024$ states are randomly sampled half-half from the human and agent validation datasets. Fig.~\ref{fig:histograms} shows histograms for activation rate $r_\Psi=\bm{\bar{\Psi}}$ of the CTR attention maps. This corresponds to the ratio of features attended to among all features.
We observe that the CTR activation rates are generally spread evenly around the targeted average one, where the agent ones feature a higher tail towards the higher values of $r_\Psi$.

Further, Fig.~\ref{fig:histograms} shows alignment scores
\begin{equation}
    a_\Psi = \frac{1}{\overline{\bm{\Psi}}} \sum_{i}^{n_\mathrm{f}} \bm{\Psi}_{i} \odot \hat{\bm{\Gamma}}_i
\end{equation}
computed between CTR attention maps $\bm{\Psi}$ and the continuous TIOA gaze prediction map $\hat{\bm{\Gamma}}$. Here, $\overline{\bm{\Psi}}$ denotes the mean value of $\bm{\Psi}$, normalizing the score such that $a_\Psi = 1$ indicates a coverage of probability mass equivalent to random attention. Values $a_\Psi > 1$ indicate better-than-random alignment, with higher values therefore reflecting stronger similarity with TIOA predictions.
We chose $a_\Psi$ over binary cross-entropy (BCE) as a tool for comparison because it provides an interpretable, activation-rate-invariant measure of relative concentration. In contrast, the expected BCE for a fixed predictor depends on the overall activation rate of $\Psi$, with higher activation generally leading to higher BCE (thus unfairly favouring the more sparse human CTR maps).

The score is computed after rounding each CTR map at $0.5$, so the maps that enter $a_\Psi$ are binary. Both human and agent maps outperform the baseline of $1.0$. Across games, the human CTR maps reach an average alignment of $3.75$ and the agent maps $3.25$ (one-sided paired $t$-test, $p=0.065$; mean difference $0.49$, 95\% CI $[-0.21,\,1.19]$, $d_z=0.74$).
The gap is largest in MsPacman. In SpaceInvaders the ordering reverses, and the agent mean ($3.98$) exceeds the human mean ($3.34$).
For Freeway, we observe a long tail of the histogram, reflecting the mode where the human is predicted to focus on the road ahead as described in Sec.~\ref{subsec:visualanalysis}. The alignment score per bin also spikes there, underlining consistency with the TIOA predictions.

To test how much of the TIOA alignment is a spatial prior, we add a center-Gaussian baseline on the $21\times 21$ grid ($\sigma=3$ cells, fixed across games). A Gaussian scaled to the CTR mean and then rounded at $0.5$ is empty at these sparsities, so the prior is instead the binary top-$k$ of that Gaussian, with $k$ matched to the number of cells active in the rounded CTR map. It is scored with the same formula on the same states. The mean center-prior alignment is $2.92$. Human CTR exceeds it in five of six games (Freeway, human $5.57$ vs.\ center $1.68$; MsPacman, human $3.21$ vs.\ center $2.21$). A one-sided paired $t$-test of human $>$ center across the six game means is not significant ($p=0.159$; mean difference $0.83$, 95\% CI $[-1.08,\,2.73]$, $d_z=0.45$). Enduro is a large exception, where the center prior reaches $5.05$, above human CTR ($3.36$), consistent with the center-focused heatmaps. An avatar-centered Gaussian was not included. Residual alignment can still reflect eye--hand coordination on the avatar.

Taken together, human CTR maps are decision-relevant features that align with overt attention beyond a simple center bias in most games. The qualitative pattern, gaze overlap plus nearby and decision-critical objects, lines up well with internal attention (overt gaze, covert attention, and short-term memory). The study has no measurement that separates this reading from action-predictive features the player did not attend.

\section{Evaluation III: Human Attention-Guided Reinforcement Learning}
\label{sec:evaluation_c}
\begin{figure*}
    \centering
    \includegraphics[width=\linewidth, clip]{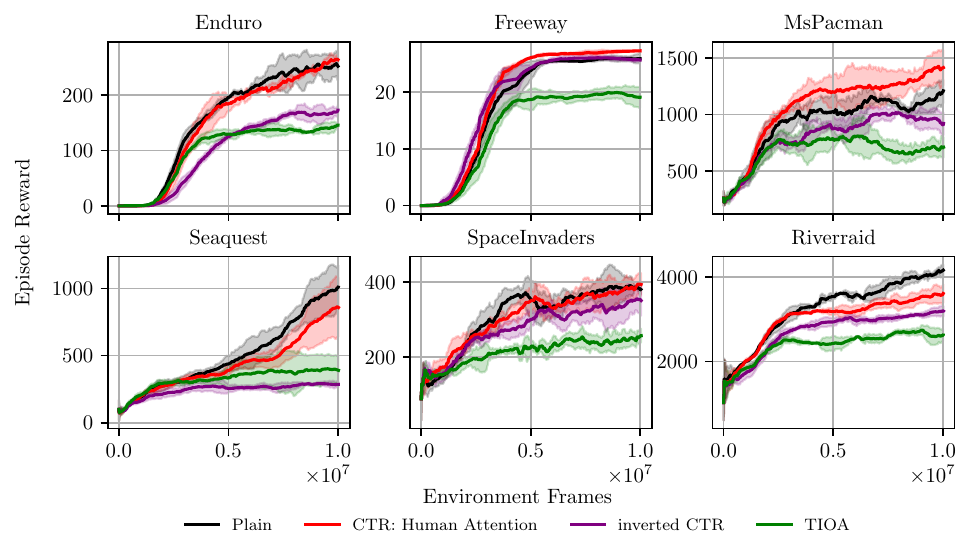}
    \caption{Comparison of learning curves of the Dueling DDQN agent with PER for the plain agent, the human attention-guided agent (CTR), the TIOA-guided agent, and the inverted attention agent. Three runs are averaged per agent type, after performing a rolling average of 200 steps per run. Shaded area indicates standard deviation. CTR-masked agents learn at a level comparable to the plain baseline, while TIOA-masked agents perform poorly.}
    \label{fig:rl_training}
\end{figure*}
Achieving human-level scores in Atari games is a challenging benchmark for RL agents, especially under limited training steps. Even algorithms such as Rainbow Deep Q-Learning (DQN)~\cite{hessel2018rainbow} that combine several improvements over normal DQN fail to reach mean human-normalized performance within 10 million environment steps, while human-outperforming methods like Agent57~\cite{badia2020agent57} are highly tailored for Atari. Guo et al.~\cite{guo2021machine} found that RL agents (here, using Proximal Policy Optimization, PPO), when they fail, are more likely to neglect regions attended to by humans, and that agent attention becomes more similar to human gaze when reducing the discount factor from 0.99 to 0.9, suggesting that human attention patterns can guide RL agents toward more human-like behavior.
We investigate whether the derived decision-relevant features are valuable and sufficient for learning, by masking RL agents to the human CTR maps and comparing them to a strong full-observation baseline, an inverted CTR mask, and a TIOA mask at matched mean activation. This is an alternative to imitation learning. We transfer the feature selection implied by human actions, rather than the actions themselves. We do not present CTR masking as a general recipe for attention-guided agents, nor do we claim a large gain over the baseline.

As a strong baseline, we use DQN with the modifications of double DQN (DDQN)~\cite{van2016deep}, dueling DQN~\cite{pmlr-v48-wangf16}, and proportional prioritized experience replay (PER)~\cite{schaul2015prioritized}. Dueling DQN improves learning stability by separating value and advantage streams, double DQN mitigates overestimation bias by using a separate target network, and PER allows the agent to learn more effectively from rare but significant events. These enhancements are implemented using a custom-modified version of the stable baselines 3 (SB3)~\cite{JMLR:v22:20-1364} DQN implementation. The policy network receives the feature representation $\mathcal{F}_\mathrm{p}$ as input. We evaluate four agent variants: (1) a plain agent using features from a pre-trained, frozen autoencoder ($\mathcal{F}_\mathrm{p} = \mathcal{F}$), (2) a human attention-guided agent, where features are masked by the CTR attention map $\bm{\Psi}$ ($\mathcal{F}_\mathrm{p} = \bm{\Psi} \odot \mathcal{F}$), with attention values below $0.1$ set to zero to prevent information leakage from low-attended features entirely and the surviving entries remain continuous weights (thresholded weighting), (3) a TIOA-guided agent, where features are masked by binarized TIOA gaze prediction maps ($\mathcal{F}_\mathrm{p} = \hat{\bm{\Gamma}}_\mathrm{bin} \odot \mathcal{F}$), and (4) an inverted attention agent, where features are masked by the inverted CTR map ($\mathcal{F}_\mathrm{p} = (1-\bm{\Psi}) \odot \mathcal{F}$). The comparison to the inverted map is done similarly to~\cite{itaya2021visual}, providing a control baseline to evaluate the usefulness of the selected features~\cite{yuezhang2018initial}. The TIOA-guided agent tests whether replacing the decision-relevant mask with a gaze top-$k$ retains learnable signal.
For each agent (1-4), we perform three runs of different seeds.
For all experiments, we use a $16\%$ mean activation rate $r_\Psi$ (average attention weight) in the CTR maps ($84\%$ mean for the inverted maps). For TIOA-guided agents, the top-$k$ binarization is set to the same $16\%$ mean activation, so the conditions are matched on mean activation while TIOA is binary and CTR uses thresholded continuous weights. Other ways to incorporate gaze (e.g., auxiliary channels or soft masking) may perform differently.

We set the Dueling DDQN and PER hyperparameters to standard values commonly used in the literature. Training is performed for 2.5 million steps (with frameskip 4, corresponding to 10 million frames), using a buffer size of 500{,}000, batch size of 32, and a learning rate of $2.5 \times 10^{-4}$. The agent starts learning after 25{,}000 steps, updates every 4 steps, and uses a target network update interval of 10{,}000. The discount factor is set to $0.99$, and exploration follows a linear schedule from $1.0$ to $0.1$ over the first 10\% of training.

Fig.~\ref{fig:rl_training} shows the learning curves for the four agent variants, denoting the reward obtained during the actual training episodes. The human CTR-guided agent sits at or above the plain agent for Enduro, Freeway and MsPacman, near it for SpaceInvaders, and below it for Seaquest and Riverraid. The inverted attention agent performs worse in most games, especially in Seaquest where almost no learning progress is made. Unintuitively, its performance in Freeway is similar to the plain agent. Where the CTR curves are visibly steady (especially Enduro and MsPacman), learning appears stable. The TIOA-masked agents perform poorly across all games, worse than the inverted CTR agents except for Seaquest.

\begin{figure}[!htb]
    \centering
    \includegraphics[width=0.5\linewidth, trim=0 0 0 0, clip]{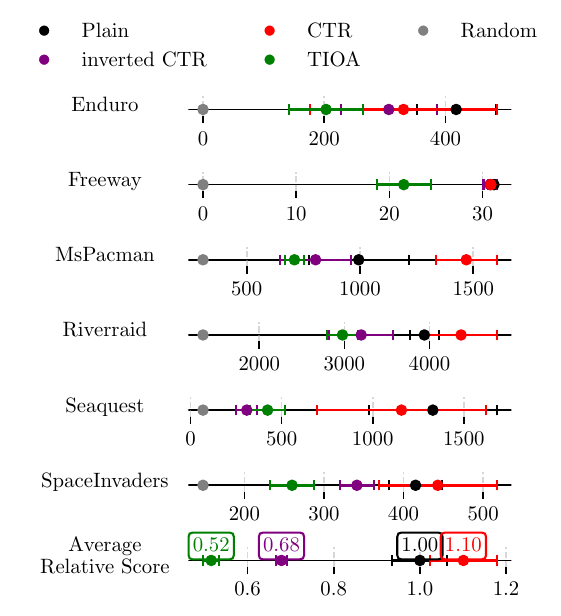}
    \caption{Comparison of episodic scores of the Dueling DDQN agent with PER for the plain agent, the human CTR-masked agent, the TIOA-masked agent, and the inverted attention agent, including the random agent score. Markers show the mean over three seeds (50 evaluation episodes each). Horizontal error bars show the standard deviation across those seeds. CTR-masked agents show a minor, non-significant average gain over the plain baseline, while TIOA-masked agents perform poorly.}
    \label{fig:episodic_score}
\end{figure}
Fig.~\ref{fig:episodic_score} shows the average episodic scores for the four agent variants after training. Here, the agents are evaluated under an exploration rate of $\epsilon=0.01$ for 150 episodes (50 per seed) per agent type, meaning they exploit their learned policies while still choosing a random action in $1\%$ of the cases. An overall average relative score is determined by
\begin{equation}
\bar{R}^{(m)} = \frac{1}{n_\mathrm{g}} \sum_{g=1}^{n_\mathrm{g}} 
\frac{S_g^{(m)} - S_g^{(\mathrm{rand})}}{S_g^{(\mathrm{plain})} - S_g^{(\mathrm{rand})}}
\end{equation}
using random scores\footnote{Random agent scores: Enduro: 0.0, Freeway: 0.0, MsPacman: 307.3, Seaquest: 68.4, SpaceInvaders: 148.0, Riverraid: 1338.5, taken from~\cite{badia2020agent57}} and plain scores as the baselines. This normalized relative score sets the plain agent to $1$ and the random agent to $0$.
The CTR-masked agent achieves highest episodic scores for MsPacman, Riverraid, and SpaceInvaders. Per-game relative scores for CTR are approximately $0.79$ (Enduro), $0.99$ (Freeway), $1.69$ (MsPacman), $1.17$ (Riverraid), $0.86$ (Seaquest), and $1.10$ (SpaceInvaders). It outperforms the inverted attention agent in all games except Freeway, where the plain, CTR-masked, and inverted agents perform similarly. The TIOA-masked agents perform poorly across all games, even below the inverted CTR-masked agent except in Seaquest. For Riverraid, the training-episode return in Fig.~\ref{fig:rl_training} (exploration on, rolling window) and the post-training evaluation in Fig.~\ref{fig:episodic_score} ($\epsilon=0.01$) disagree in ordering relative to the plain agent. We therefore report both and do not smooth them into a single claim.

For the average relative score, the CTR-masked agents achieve $1.10$, while the inverted attention agents underperform with a score of $0.68$, and the TIOA-masked agents achieve $0.52$. With only $16\%$ mean activation, this is a minor improvement over the full-observation Dueling DDQN+PER agent and is not statistically significant. One-sided paired $t$-tests across the six games (game means are themselves means of three seeds) give CTR vs.\ plain $p=0.238$ (mean relative-score difference $0.10$, 95\% CI $[-0.24,\,0.44]$, $d_z=0.31$), CTR vs.\ inverted $p=0.018$ (difference $0.42$, 95\% CI $[0.04,\,0.81]$, $d_z=1.15$), and CTR vs.\ TIOA $p=0.0024$ (difference $0.59$, 95\% CI $[0.27,\,0.90]$, $d_z=1.97$). The experiment shows that the selected features are sufficient for an agent to learn at that baseline. Raising performance above the plain agent would require exploration of other approaches, such as soft masking techniques or approaches that align the agent's attention with the human CTR maps. The inverted and TIOA contrasts show that these features, specifically, carry the learnable signal.

The lower performance of TIOA-masked agents shows that a gaze top-$k$ at matched mean activation does not retain the same learnable signal as the action-trained CTR mask in this hard feature-selection setup.

\section{Conclusions}
\label{sec:discussion}
Our study presents a method to derive decision-relevant visual features from gameplay data alone, leveraging offline attention methods from reinforcement learning (RL). The contextualized, task-relevant (CTR) attention network generates attention maps from human and agent gameplay. We compare them to a temporally-integrated overt attention (TIOA) gaze pipeline and to a center-prior baseline, and we apply the human maps to attention-guided RL.

Evaluation across six Atari games shows that human CTR maps identify features sufficient for action prediction and are sparser than the matched agent maps. Their mean TIOA alignment ($3.75$) is higher than the agent maps ($3.25$, paired $p=0.065$) and exceeds a binary center prior ($2.92$) in five of six games (Enduro is the center-bias exception, paired human $>$ center $p=0.159$). Qualitatively, the maps capture key TIOA regions together with nearby and highly decision-relevant objects. This pattern lines up well with internal attention (overt gaze, plus covert attention and short-term memory) as an interpretation of the maps, while remaining open to the alternative that some selected features were never attended.

Applied to RL, CTR-masked agents reach a normalized relative score of $1.10$ against the plain Dueling DDQN+PER baseline at $16\%$ mean activation, a minor improvement that is not statistically significant (paired $p=0.238$), and significantly outperform TIOA-masked and inverted CTR-masked agents. The selected features are thus sufficient for an agent to learn at that baseline. Raising performance above it may need exploration of approaches other than the hard mask, such as soft masking or aligning the agent's attention with the human maps.

Limitations include the need for larger human gameplay datasets with eye-tracking and the random transition split, since inputs use frame stacks and the gaze target is integrated over time. The analysis pools four subjects. Action prediction and TIOA alignment both come from the same gameplay, and there is no external validation of the covert-attention reading. We also do not compare CTR directly with other saliency or self-supervised attention extractors beyond the positioning in the introduction. Methods could further be extended to datasets from autonomous driving or robotics.

These results confirm that the method successfully derives decision-relevant features from gameplay alone, without brain-activity recordings. They also provide a basis for discussing how those features relate to human internal attention. Future work could test that discussion with independent validation and explore further uses of such features in explainable and human-inspired agents.

\backmatter

\bmhead{Supplementary information}
We provide supplementary animations of the CTR and TIOA attention maps, namely the binarized maps shown in Fig.~\ref{fig:att_comparison}, the smoothed maps shown in Fig.~\ref{fig:att_comparison_smooth}, as well as the maps binarized at average activation of 16\% as used in the RL experiments in Sec.~\ref{sec:evaluation_c}.

\section*{Declarations}
\noindent\textbf{Funding:} The authors did not receive support from any organization for the submitted work.\\
\noindent\textbf{Competing Interests:} The authors declare no competing interests.\\
\noindent\textbf{Data availability:} Source code to create the results of this study is available at \url{https://github.com/UThenrik/revealing-attention}.\\
\noindent\textbf{Author contribution:} H.K. carried out the research and wrote the manuscript. T.Y. supervised the study and provided guidance. All authors reviewed the manuscript.




\bibliography{references}

\end{document}